\documentclass{article}

\PassOptionsToPackage{numbers, compress}{natbib}

\usepackage[preprint]{neurips_2022}

\usepackage[utf8]{inputenc} 
\usepackage[T1]{fontenc}    
\usepackage{hyperref}       
\usepackage{url}            
\usepackage{booktabs}       
\usepackage{amsfonts}       
\usepackage{nicefrac}       
\usepackage{microtype}      
\usepackage{xcolor}         
\usepackage{graphicx}
\usepackage{enumerate}
\usepackage{amsmath}
\usepackage{multirow}
\usepackage{subfigure}
\usepackage{wrapfig}
\usepackage[ruled,linesnumbered]{algorithm2e}
\title{MemoNav: Selecting Informative Memories for Visual Navigation}

\author{
Hongxin Li \\
  CASIA \\
  \texttt{lihongxin2021@ia.ac.cn} \\
  \And
  Xu Yang \\
  CASIA \\
  \texttt{xu.yang@ia.ac.cn} \\
  \AND
  Yuran Yang \\
  Tencent Map, T Lab \\
  \texttt{yuranyang@tencent.com} \\
  \And
  Shuqi Mei \\
  Tencent Map, T Lab \\
  \texttt{shawnmei@tencent.com} \\
  \And
  Zhaoxiang Zhang \\
  CASIA \\
  \texttt{zhaoxiang.zhang@ia.ac.cn} \\
}

\begin{document}

\maketitle

\begin{abstract}
  Image-goal navigation is a challenging task, as it requires the agent to navigate to a target indicated by an image in a previously unseen scene. Current methods introduce diverse memory mechanisms which save navigation history to solve this task. However, these methods use all observations in the memory for generating navigation actions without considering which fraction of this memory is informative. To address this limitation, we present the MemoNav, a novel memory mechanism for image-goal navigation, which retains the agent's informative short-term memory and long-term memory to improve the navigation performance on a multi-goal task. The node features on the agent's topological map are stored in the short-term memory, as these features are dynamically updated. To aid the short-term memory, we also generate long-term memory by continuously aggregating the short-term memory via a graph attention module. The MemoNav retains the informative fraction of the short-term memory via a forgetting module based on a Transformer decoder and then incorporates this retained short-term memory and the long-term memory into working memory. Lastly, the agent uses the working memory for action generation. We evaluate our model on a new multi-goal navigation dataset. The experimental results show that the MemoNav outperforms the SoTA methods by a large margin with a smaller fraction of navigation history. The results also empirically show that our model is less likely to be trapped in a deadlock, which further validates that the MemoNav improves the agent's navigation efficiency by reducing redundant steps.
\end{abstract}

\section{Introduction}
Human beings plan their routes by recalling detailed experiences of the past \cite{tulving1985memory, suddendorf2009mental, manning2021episodic}. When we get into a new environment, the first thing is usually to explore our surroundings to get familiar with the scene layout. This exploration behavior enables us to quickly plan new routes with the help of the so-called \emph{working memory} \cite{1960Plans,cowan2008differences}.\par
Recent neuroscience research \cite{ravizza2021working, mansouri2015working, blacker2017keeping} has found that working memory is essential for organizing goal-directed behaviors including navigation, as it maintains task-relevant information. Several models \cite{ericsson1995long, cowan2008differences, baddeley2012working} of working memory have been constructed to explain the relationships among short-term memory (STM), long-term memory (LTM), and working memory (WM). Particularly, the model by Cowan et al. \cite{cowan2008differences} (shown in the Appendix) demonstrates that the human brain selects short-term memory as part of working memory by focusing attention on task-relevant experiences to avoid distractions. It is also noticed that working memory is not a simple extension of short-term memory, but an incorporation of selected short-term memory and long-term memory in goal-oriented tasks \cite{baddeley2012working}. \par
Inspired by the above neuroscience research, we propose a MemoNav for image-goal visual navigation (ImageNav) agents, which is built upon a SoTA topological map-based method VGM \cite{kwon2021visual}. The MemoNav simulates the working memory mechanism of the human brain and improves the agent's ability to complete multi-goal navigation tasks. Recently, a number of ImageNav methods \cite{kwon2021visual, chaplot2020neural, chaplot2020learning, fang2019scene, kumar2018visual} have introduced memory mechanisms. According to the structure of memory, these methods can be classified into three categories: (a) metric map-based methods \cite{chaplot2020learning, chen2018learning} that reconstruct local top-down maps and aggregate them into a global map, (b) stacked memory-based methods \cite{pashevich2021episodic, mezghani2021memory, fang2019scene} that stack the past observations and pose sensor data in chronological order, and (c) topological map-based methods \cite{kwon2021visual, chaplot2020neural, beeching2020learning, savinov2018semi} that store sparse landmark features in the graph nodes. Comparatively speaking, the metric map-based methods store every detail of the scene, thus consuming large memory when the agent encounters large scenes; the stacked memory-based methods utilize all past data without differentiating useful information and noise; the topological map-based methods benefit from the memory sparsity of topological maps, but still suffer from all memory computation without considering the contribution of each graph node.\par

To overcome these limitations, the MemoNav enables the agent to flexibly retain informative navigation experience. In our design, the MemoNav utilizes the aforementioned three types of memory: \textbf{STM}, \textbf{LTM}, and \textbf{WM}. STM represents the local node features in a topological map; LTM is a global node continuously aggregating short-term memory; WM incorporates the informative fraction of the STM and the LTM. Technically, the MemoNav consists of four steps: map update, selective forgetting, memory encoding and decoding, and action generation. (1) Firstly, the map update module stores landmark features on the map as the STM. (2) To incorporate informative STM into the WM, a selective forgetting module temporarily removes nodes whose attention scores assigned by the memory decoder rank below a predefined percentage. After forgetting, the navigation pipeline will not compute the forgotten node features at subsequent time steps. To assist the STM, we add a global node \cite{ainslie2020etc,beltagy2020longformer} to the map as the LTM. The global node links to all map nodes and continuously aggregates the features of these nodes at each time step. (3) The retained STM and the LTM form the WM, which is then processed by a GATv2 \cite{brody2022how} encoder. Afterward, two Transformer \cite{vaswani2017attention} decoders use the embeddings of the goal image and the current observation to decode the processed WM. (4) Lastly, the decoded features are used for generating navigation actions.\par
The MemoNav enables the agent to utilize task-relevant experiences to find a short path to the goal without being intervened by misleading information. With the synergy of these three types of memory, the MemoNav achieves considerable progress in multi-goal ImageNav tasks. The results show that compared with the SoTA method \cite{kwon2021visual}, the MemoNav increased the success rate by 5\%, 9\%, and 3\% on 2-goal, 3-goal, and 4-goal test datasets, respectively.
The main contributions of this paper are as follows:
\begin{itemize}
\item We propose the MemoNav, which utilizes a forgetting module to select informative short-term memory stored in the topological map for generating navigation actions.
\item We introduce a global node to represent long-term memory as a supplement to the short-term memory.
\item We evaluate the proposed model on a multi-goal task. The results demonstrate that our model outperforms the SoTA baseline by a large margin on a multi-goal task.
\end{itemize}


\section{Related Work}
\textbf{ImageNav models based on topological maps}. Cognitive research \cite{wang2002human, foo2005humans, gillner1998navigation} suggests that humans save landmark features in their memory for navigation, instead of detailed scene layouts. Several methods \cite{savinov2018semi,beeching2020learning,chaplot2020neural,kwon2021visual,chen2021topological} utilize this theory and propose topological memory for visual navigation. SPTM \cite{savinov2018semi} is an early attempt to use topological maps in ImageNav. It proposes a semi-parametric topological memory that is pre-built before navigation. Beeching et al. \cite{beeching2020learning} introduced a graph processing method inspired by dynamic programming-based shortest path algorithms. This method also needs to explore the scene in advance. NTS \cite{chaplot2020neural} and VGM \cite{kwon2021visual} incrementally build the topological map during navigation. VGM uses no pose sensor data and is more robust to sensor noise than NTS. The two methods utilize all features in the map, while the MemoNav flexibly utilizes the informative fraction of these features.\par
\textbf{Memory mechanisms for reinforcement learning}. 
Several studies \cite{ritter2021rapid, lampinen2021towards, sukhbaatar2021not, loynd2020working} draw inspiration from memory mechanisms of the human brain and design reinforcement learning models for reasoning over long time horizons. Ritter et al. \cite{ritter2021rapid} proposed an episodic memory storing state transitions for navigation tasks. Lampinen et al. \cite{lampinen2021towards} presented hierarchical attention memory as a form of ``mental time-travel'' \cite{tulving1985memory}, which means recovering goal-oriented information from past experiences. Unlike this method, our model retains such information via a novel forgetting module. Expire-span \cite{sukhbaatar2021not} predicts life spans for each memory fragment and permanently deletes expired ones. Our model is different from it in that we restore forgotten memory if the agent returns to visited places. The most similar work is WMG \cite{loynd2020working} which also uses theories of working memory. However, its memory capacity is fixed. In contrast, our model retains a certain proportion of short-term memory in the working memory and adjusts the memory capacity when the navigation episode contains more goals.

\section{Background}
\subsection{Task Definition}
\label{sec: task def}
\begin{figure}
  \centering
  \includegraphics[width=0.7\linewidth]{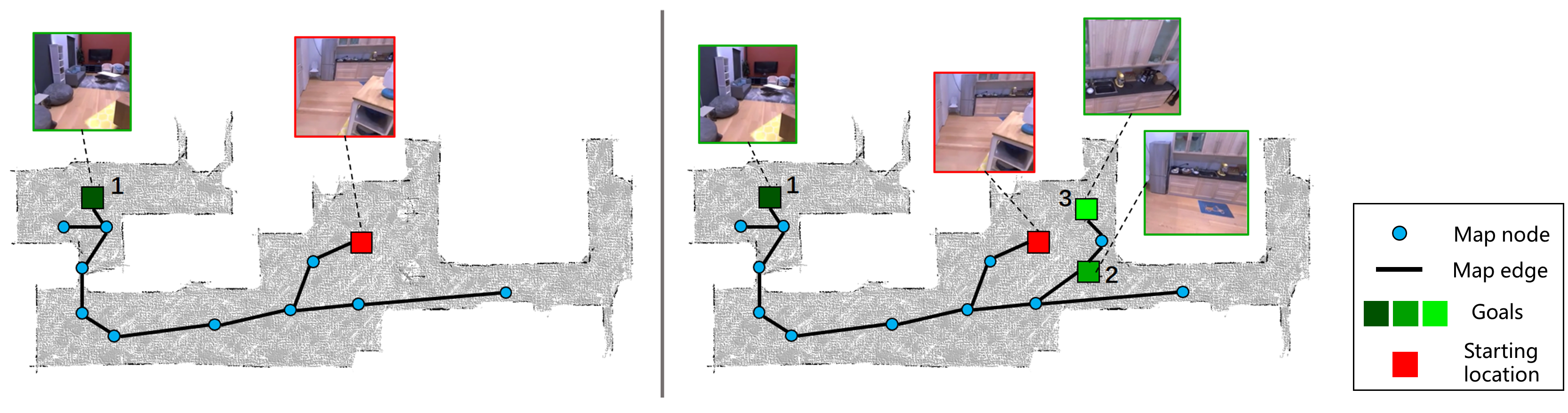}
  \caption{\textbf{Example episodes for 1-goal and three-goal tasks.} In the 1-goal task (left), the agent simply explores the scene, finds a path leading to the destination. In the three-goal task (right), the agent continues to navigate to two more successive goals that are probably set in visited areas.}
  \label{fig:task_comparison}
\end{figure}

We consider image-goal navigation (ImageNav) in this paper. As shown in Figure~\ref{fig:task_comparison} (left), the objective of the agent is to learn a policy $\pi(a_t|s_t, \mathbf{I}_{target})$ to reach a target, given an image $\mathbf{I}_{target}$ that contains a view of the target and a series of observations, $\left\{\mathbf{I}_t \right\}$, captured during the navigation. At the beginning of navigation, the agent receives an RGB image, $\mathbf{I}_{target}$, of the target. At each time step, it captures an RGB-D panoramic image, $\mathbf{I}_t$ of the current location and generates a navigational action. Following \cite{kwon2021visual}, any additional sensory data, e.g. GPS and IMU, are not available. \par

\subsection{Brief Review of Visual Graph Memory}
\label{sec: VGM}
The MemoNav is mainly built upon VGM \cite{kwon2021visual}, which is briefly introduced below.\par
VGM incrementally builds a topological map $G=\left\{V, E\right\}$ from the agent's past observations where $V$ and $E$ denote nodes and edges, respectively. The node features (STM) $\mathbf{V}\in \mathbb{R}^{d\times N_{t}}$ are generated from observations by a pretrained encoder $\mathcal{F}_{loc}$ where $d$ denotes the dimension of feature and $N_{t}$ the number of nodes at time $t$. \par
VGM updates the map in two steps: localization and graph update. \textbf {Localization} is implemented by calculating embedding similarity. VGM first uses $\mathcal{F}_{loc}$ to map the current observation $\rm{I}_\emph{t}$ to an embedding $e_t\in \mathbb{R}^d$, and then calculates the similarities $\left\{s_{i} \mid s_{i}=\frac{n_{i} \cdot e_{t}}{\left\|n_{i}\right\|\left\|e_{t}\right\|}, i=1,2, \ldots, N_{t}\right\}$ between $e_t$ and each node feature $\mathbf{V}=\left\{n_1, n_2, ..., n_{N_t}\right\}$. If the similarity of a node $s_i$ is higher than a threshold $s_{th}$, this node is seen as near the agent. \textbf {Graph Update} depends on three cases. (i) If the agent is localized at the last node, the map will not be updated. (ii) If the localized node is different from the last one, a new edge connecting these two nodes will be created, and the feature of the localized node will be replaced with the new embedding. (iii) If the agent fails to be localized, a new node with the current observation embedding will be created, and a new edge will connect this new node with the last node. \par
VGM utilizes a three-layer graph convolutional network (GCN) to process the topological map and obtain the encoded memory $\mathbf{M}$ (i.e., a sequence of feature vectors). Before the first layer, VGM fuses each feature of a node with the target image embedding $e_{target}=\mathcal{F}_{enc}(\mathbf{I}_{target})$ using a linear layer. \par 



The encoded memory is then decoded by two Transformer decoders, $\mathcal Dec_{cur}$ and $\mathcal Dec_{target}$. $\mathcal Dec_{cur}$ takes the current observation embedding $e_{cur}=\mathcal{F}_{enc}(\mathbf{I}_{cur})$ as the query and the feature vectors of the encoded memory $\mathbf{M}$ as the keys and values, generating a feature vector $f_{cur}$. Similarly, $\mathcal Dec_{target}$ takes the target embedding $e_{target}$ as the query and generates $f_{target}$.
Lastly, an LSTM-based policy network takes as input the concatenation of $f_{cur}$, $f_{target}$ and $e_{cur}$, and outputs an action distribution.

\section{Method}
In this section, We expand the definition of ImageNav and describe our MemoNav designed for multi-goal navigation tasks. 
\subsection{Multi-goal ImageNav}
Single-goal tasks are not sufficient for thoroughly evaluating the potential of memory mechanisms, since finding a single goal may not need all navigation history. Hence, it remains unclear which fraction of navigation history is useful for finding the goal. To further investigate how memory mechanisms assist the agent in exploring the environment and locating the target, we borrow the idea of MultiON \cite{wani2020multion} and collect multi-goal test datasets (an example in Figure~\ref{fig:task_comparison}). In a multi-goal task, the agent navigates to an ordered sequence of $N$ destinations $\left\{ \rm {T}\right\}_1^N$ in the environment. We place the final target within the vicinity of previous ones to form a loop. By letting the agent return to visited places, we are able to test whether memory mechanisms can help the agent plan a short path. If not, the agent will probably waste its time re-exploring the scene or traveling randomly.

\begin{figure}
  \centering
  \includegraphics[width=0.8\linewidth]{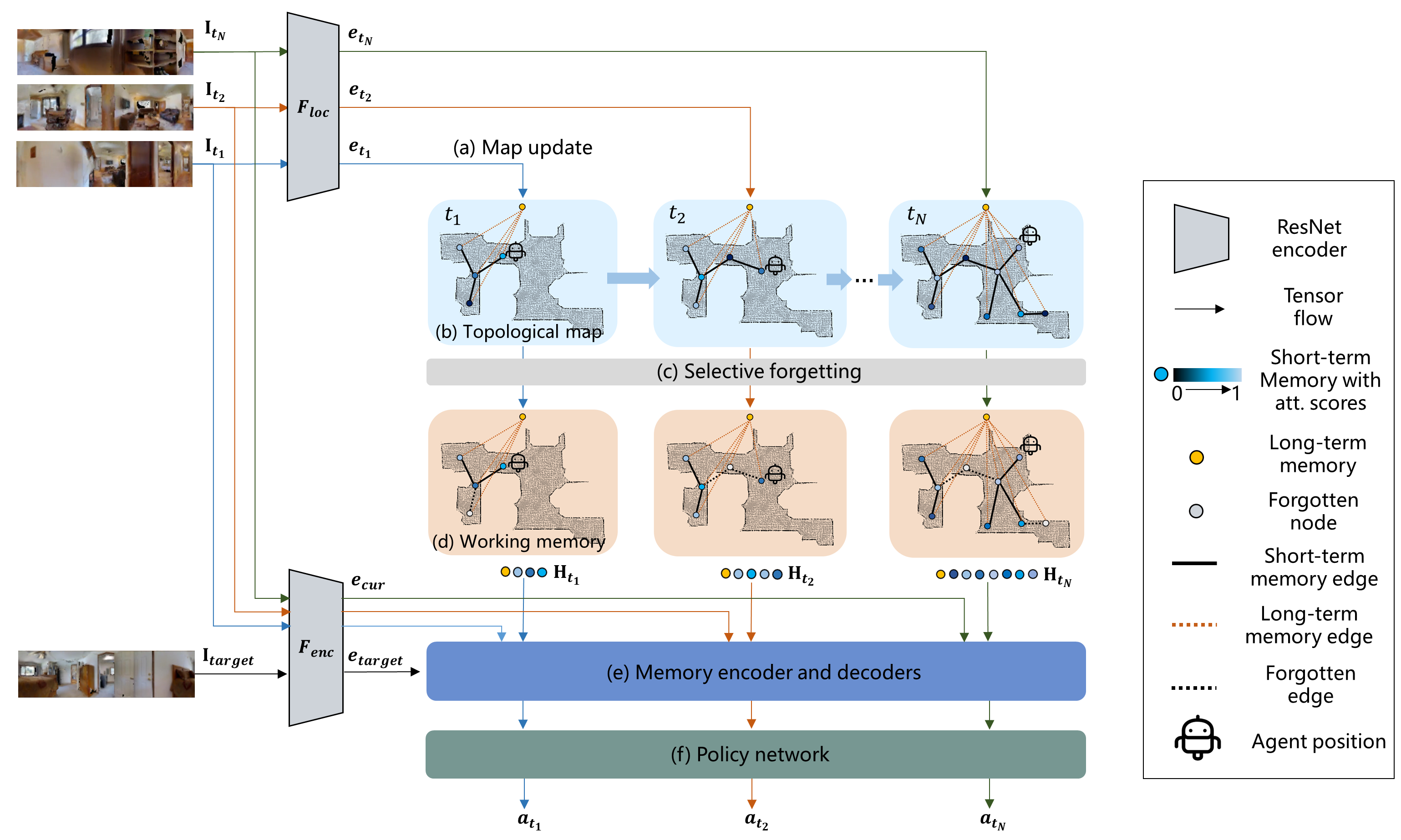}
  \caption{\textbf{Overview of the MemoNav.} (a) The memory update module builds (b) a topological map using $e_t$, the embedding of the current image $\mathbf{I}_t$. The node features in the map form the short-term memory while a global node that links to each of these map nodes acts as the long-term memory. (c) The forgetting module temporarily excludes a fraction of short-term memory whose attention scores rank below a threshold $p$. (d) The working memory $\mathbf{H}_t$ incorporates the retained fraction of the short-term memory and long-term memory for (e) the subsequent memory encoding and decoding (details in the Appendix). Lastly, the output of the decoding process is input to (f) the policy network.}
  \label{fig:model_overview}
\end{figure}

\subsection{Model Components}
\label{sec: components}
The MemoNav comprises three key components: the forgetting module, the global node, and the GATv2-based memory encoder. We show the pipeline of the MemoNav in Figure~\ref{fig:model_overview} and describe these components in the remainder of this section.

\subsubsection{Working Memory Mechanism via Selective Forgetting }
\label{sec: forgetting mechanism}
\begin{figure}
  \centering
  \includegraphics[width=0.7\linewidth]{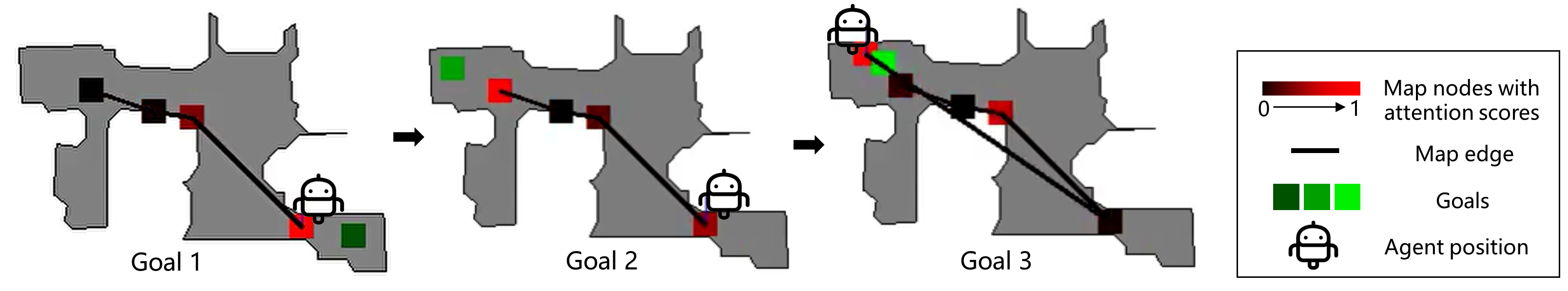}
  \caption{\textbf{VGM calculates redundant node features}. The figure visualizes the attention scores in $\mathcal Dec_{target}$ for an example episode.}  
  \label{fig:redundancy}
\end{figure}

VGM (Section~\ref{sec: VGM}) decodes all STM in the map, thus introducing redundancy. Inspired by previous studies on the forgetting mechanisms of animal brains \cite{manning2021episodic, nematzadeh2020memory, sabandal2021dopamine, suddendorf2009mental, tulving1985memory}, we devise a forgetting module that instructs the agent to forget uninformative experiences. Here, `forgetting' means that nodes with attention scores lower than a threshold are temporarily excluded from the navigation pipeline. This means of forgetting via attention is evidenced by research \cite{anderson2018evaluation, fukuda2009human} revealing that the optimal performance of working memory relies on humans' ability to focus attention on task-relevant information and to avoid distractions. In Figure~\ref{fig:redundancy}, we visualize the attention scores calculated in $\mathcal Dec_{target}$ for the encoded STM. The figure shows that the agent assigns high attention scores to nodes close to targets while paying little to no attention to remote ones. This phenomenon indicates that not all node features in the STM are helpful and that it is more efficient to incorporate a small number of them in the WM for navigation. \par
With these insights, we propose a selective forgetting module that helps to select STM to form WM according to the attention scores $\left\{\alpha_{i}\right\}_{i=1}^{N_t}$ the goal embedding $e_{target}$ assigns to all nodes in $\mathcal Dec_{target}$. Each time $\mathcal Dec_{target}$ finishes decoding, the agent temporarily `forgets' a fraction of nodes whose scores rank below a predefined percentage $p$. In other words, these nodes will be disconnected from their neighbors, and not be processed by the navigation pipeline at the subsequent time steps. If the agent returns to a forgotten node, this node will be added to the map and processed by the pipeline again. In a multi-goal task, once the agent has reached a target, all forgotten nodes will be restored, as the nodes that are uninformative for finding the last target may lead the agent to the next one. \par
The proposed forgetting mechanism is used in a plug-and-play manner. Our default model sets $p$ as 20\%. We train our model without forgetting in order for it to learn to assign high attention scores to the informative fraction of STM. When evaluating and deploying the agent, we switch on the forgetting mechanism. With this mechanism, the agent can selectively incorporate the most informative STM in the WM, while avoiding misleading experiences.
\subsubsection{Long-term Memory Generation}
\label{sec: Global node}
In addition to STM, knowledge from LTM also forms part of WM \cite{ericsson1995long}. Inspired by ETC \cite{ainslie2020etc} and LongFormer \cite{beltagy2020longformer}, we add a zero-initialized global node $n_{global} \in \mathbb{R}^d$ to the topological map as the LTM. This global node is respawned at the beginning of each episode and continuously fuses the STM via memory encoding (Figure~\ref{fig:model_overview} (e)), thereby storing past observations of the agent. Moreover, the global node acts as the LTM in that it contains a scene-level feature. A recent study \cite{ramakrishnan2022environment} suggests that embodied agents must learn higher-level representations of the 3D environment. From this viewpoint, the LTM stores a high-level spatial representation of the scene by aggregating local node features.\par
Another merit of the LTM is facilitating feature fusing. The topological map is divided into several sub-graphs when forgotten nodes are removed. Consequently, direct message passing between the nodes separated in different sub-graphs no longer exists. The global node links to every node in the map, thereby acting as a bypath that facilitates feature fusing between these isolated sub-graphs.\par

\subsubsection{GATv2-based Memory Processor}
To generate LTM from STM that is dynamically updated by the agent, we change the backbone of the memory encoder in VGM (Section~\ref{sec: VGM}) from GCN to GATv2 \cite{brody2022how}. GATv2 differs from GCN \cite{kipf2017semi} in that the weights for neighboring nodes are derived from node features, instead of the Laplacian matrix. \par
Brody et al. \cite{brody2022how} claimed that GATv2 is powerful when different nodes have different rankings of neighbors. This is exactly the case in multi-goal ImageNav: the extents to which different nodes contribute to leading the agent to the goal are diverse. When the LTM aggregates the STM via the attention mechanism in GATv2, the STM features that contain information about the target or a path leading to the target should obtain high attention, while those of irrelevant places should receive lower attention. Furthermore, the node features in the STM are dynamically updated during navigation, as the agent replaces these features with new ones when revisiting these nodes. Therefore, the dynamic attention mechanism of GATv2 is more suitable for processing the STM that dynamically varies during navigation. \par
Combining the STM $\mathbf{V}$, LTM $n_{global}$ and GATv2-based processor $\mathcal{F}_{proc}$, the memory processing is formulated as follows:
\begin{equation}
\left\{\mathbf{V}^{l+1}, n_{global}^{l+1} \right\}=\operatorname{GATv2}\left(\left\{\mathbf{V}^{l}, n_{global}^{l} \right\} \right), \quad 
\mathbf{M}^{\prime}=\mathcal{F}_{proc}(G; \psi) = \left\{\mathbf{V}^{3}, n_{global}^{3} \right\}
\end{equation}
where $l=0,1,2$ represents the layer number, $\mathbf{M}^{\prime}$ the encoded working memory, and $\psi$ the processor parameters. $\left\{\cdot, \cdot\right\}$ denotes concatenation along the sequence dimension. \par
After the processing of GATv2, the decoders $\mathcal Dec_{cur}$ and $\mathcal Dec_{target}$ take $\mathbf{M}^{\prime}$ as keys and queries, generating $f_{target}$ and $f_{cur}$, which fuse both LTM and STM.

\section{Experiments}
\subsection{Simulation Settings}
All experiments are conducted in the Habitat \cite{savva2019habitat} simulator with the Gibson \cite{xia2018gibson} scene dataset. Following VGM \cite{kwon2021visual}, we train all models with 72 scenes, and evaluate them on a public 1-goal dataset \cite{mezghani2021memory} that comprises 1400 episodes for 14 unseen scenes. For a fair comparison, 1007 sampled episodes are used for evaluating our model on 1-goal tasks, while 1400 episodes are used for conducting ablation studies. In addition, we generate 700-episode datasets for a multi-goal task by ourselves (see the Appendix). The difficulty of an episode is indicated by the number of goals \footnote{The 1-goal difficulty level here denotes the hard level in the public test dataset}  All baselines and our model are trained on the 1-goal dataset and tested on other difficulty levels.\par
The action space of the agent consists of four options: \verb+stop+, \verb+forward+, \verb+turn_left+, and \verb+turn_right+. When the agent performs a \verb+forward+ action, the agent moves 0.25m forward. A turning action rotates the agent by 10 degrees. The agent is allowed to take at most 500 steps in either 1-goal or multi-goal tasks. The agent succeeds in an episode if it performs \verb+stop+ within a 1m radius of each target location. The agent fails if it performs \verb+stop+ anywhere else or runs out of time step budget.

\subsection{Training Techniques}
Following VGM \cite{kwon2021visual}, we first train the agent using imitation learning, minimizing the negative log-likelihood of the ground-truth actions. Next, we finetune the agent’s policy with proximal policy optimization (PPO) \cite{schulman2017proximal} to improve the exploratory ability of the agent. The reward setting and auxiliary losses remain the same as in VGM. All our models are obtained on a TITAN RTX GPU.


\subsection{Baselines}

We compare the proposed model with baselines that also utilize memory mechanisms. \textbf{Random} is an agent taking random actions with oracle stopping. \textbf{ANS} \cite{chaplot2020learning} and \textbf{Exp4nav} \cite{chen2018learning} are metric map-based models proposed for the task of exploration. They are adapted for ImageNav in the experiments of VGM \cite{kwon2021visual}. \textbf{SMT} \cite{fang2019scene} stacks past observations in chronological order and uses a Transformer decoder to further process them. \textbf{Neural Planner} \cite{beeching2020learning} needs to explore the scene first to pre-build a topological map. Once the map is built, this method performs a hierarchical planning strategy to generate navigation actions. \textbf{NTS} \cite{chaplot2020neural} incrementally builds a topological map without pre-exploring. Its navigation strategy follows two steps: a global policy samples a subgoal, and then a local policy takes navigational actions to reach the subgoal. \textbf{VGM} \cite{kwon2021visual} (see Section \ref{sec: VGM}) is the baseline the MemoNav is built on. The quantitative results of these baselines on 1-goal tasks are from \cite{kwon2021visual}. We re-evaluate VGM and report the new results.

\subsection{Evaluation Metrics}
\label{sec: eval metrics}
In 1-goal tasks, the success rate (\textbf{SR}) and success weighted by path length (\textbf{SPL}) \cite{anderson2018evaluation} are used. In a multi-goal task, we borrow two metrics from \cite{wani2020multion}: the Progress (\textbf{PR}) and progress weighted by path length (\textbf{PPL}). \textbf{PR} is the fraction of goals that are successfully reached, equal to the \textbf{SR} for 1-goal tasks. \textbf{PPL} indicates navigation efficiency and is defined as $PPL=\frac{1}{E} \sum_{i=1}^{E} Progress_{i} \frac{l_{i}}{\max \left(p_{i}, l_{i}\right)}$, where $E$ is the total number of test episodes, $l_i$ and $p_i$ are the shortest path distance to the final goal via midway ones, and the actual path length taken by the agent, respectively.

\subsection{Quantitative Results}


\begin{table}[]
\scriptsize
\caption{Evaluation results of the baselines and our model on the 1-goal test dataset. (\textbf{SR}: success rate, \textbf{SPL}: success weighted by path length)}
\label{tab: single_goal}
\centering
\begin{tabular}{@{}cccc|cccc@{}}
\toprule
Methods & Memory Type & SR & SPL & Methods        & Memory Type & SR & SPL \\ \midrule
Random  &     -       & 0.17  & 0.05  & Neural Planner \cite{beeching2020learning} &   graph    & 0.42 & 0.27 \\
ANS \cite{chaplot2020learning} &  metric map   & 0.30 &  0.11 & NTS \cite{chaplot2020neural}           &   graph       & 0.43 & 0.26  \\
Exp4nav \cite{chen2018learning} &  metric map & 0.47 &  0.39 & VGM \cite{kwon2021visual}            &    graph    & 0.75  & \textbf{0.58}  \\
SMT  \cite{fang2019scene} &    stack    & 0.56 &  0.40 & MemoNav (ours) &     graph    & \textbf{0.78} & 0.54  \\ \bottomrule
\end{tabular}
\end{table}


\begin{table}[]
\scriptsize
\caption{Evaluation results of the VGM and our model on the multi-goal test datasets. (\textbf{PR}: progress, \textbf{PPL}: progress weighted by path length)}
  \label{tab: multi goal}
  \centering
\begin{tabular}{@{}ccccccc@{}}
\toprule
\multirow{2}{*}{Methods}  & \multicolumn{2}{c}{\textbf{2-goal}} & \multicolumn{2}{c}{\textbf{3-goal}} & \multicolumn{2}{c}{\textbf{4-goal}} \\ \cmidrule(lr){2-3} \cmidrule(lr){4-5} \cmidrule(lr){6-7} 
 &  PR & PPL & PR & PPL & PR & PPL \\ \midrule
Random &  0.05 & 0.01 & 0.03 & 0.00 & 0.02 & 0.00 \\
VGM \cite{kwon2021visual} &  0.45 & \textbf{0.18} & 0.33 & 0.08 & 0.28 & 0.05 \\
MemoNav (ours)  & \textbf{0.50} & 0.17 & \textbf{0.42} & \textbf{0.09} & \textbf{0.31} & \textbf{0.05} \\
\bottomrule
\end{tabular}
\end{table}

\textbf{Comparison on 1-goal tasks}. Table~\ref{tab: single_goal} shows that the proposed method outperforms all the baselines in SR on the 1-goal dataset. Compared with the metric map-based methods (Exp4nav \cite{chen2018learning} and Neural Planner \cite{beeching2020learning}), our model enjoys a large improvement in SR (from \textbf{0.47} to \textbf{0.78}) and SPL (from \textbf{0.39} to \textbf{0.54}). The two baselines require pre-built maps but exhibit lower SR and SPL. This is partly because the explored areas fail to cover the target place. Compared with VGM \cite{kwon2021visual}, our model exhibits a slight performance gain, increasing the SR by \textbf{0.03} while using 20\% fewer node features. This result indicates that VGM uses redundant past observations that probably interfere with the agent's navigation. In contrast, our agent uses the informative fraction of these observation, obtaining a higher success rate with less past information. The SPL slightly decreases, probably because the agent forgets a certain proportion of explored places and spends more time re-exploring.

\textbf{Comparison on a multi-goal task}. In Table~\ref{tab: multi goal}, we compare the MemoNav with VGM \cite{kwon2021visual} on a multi-goal task. As the difficulty level rises from 1 goal to 4 goals, both methods witness drops in the navigation success rate, but the MemoNav is more robust and more competent to complete a multi-goal task. At the 2-goal level, the MemoNav outperforms VGM in PR by \textbf{0.05} (\textbf{11.1\%}). When the number of goals increases to 3, the performance gap widens: the improvement in PR rises to \textbf{0.09} (\textbf{27.2\%}). At the 4-goal level, the two methods both perform worse, but our method still surpasses VGM in PR by a noticeable margin of \textbf{0.03} (\textbf{10.7\%}).

\subsection{Ablation Studies and Analysis}
\label{sec: ablation}

\begin{table}[]
\scriptsize 
\caption{Network component ablation results. Row 1 is the baseline model VGM \cite{kwon2021visual}, and row 7 is our full model. (\textbf{GATv2}: the GATv2-based memory encoder, \textbf{LTM}: the long-term memory, \textbf{FG20\%}: forgetting nodes whose attention scores rank below 20\%) }
  \label{tab: component ablation}
\centering
\begin{tabular}{@{}cccccccccccc@{}}
\toprule
\multirow{2}{*}{} & \multicolumn{3}{c}{\textbf{Components}} & \multicolumn{2}{c}{\textbf{1-goal}} & \multicolumn{2}{c}{\textbf{2-goal}} & \multicolumn{2}{c}{\textbf{3-goal}} & \multicolumn{2}{c}{\textbf{4-goal}} \\ \cmidrule(lr){2-4} \cmidrule(lr){5-6} \cmidrule(lr){7-8} \cmidrule(lr){9-10}  \cmidrule(lr){11-12}
  & GATv2 & LTM &FG20\% & SR & SPL & PR & PPL & PR & PPL & PR & PPL \\ \midrule
1 &          &         &           & 0.621 & 0.494 & 0.449 & 0.178 & 0.329 & 0.080 & 0.276 & 0.054 \\
2 & \checkmark   &         &           & 0.618 & \textbf{0.533} & 0.452 & 0.193 & 0.377 & 0.080 & 0.307 & 0.060 \\
3 &          & \checkmark   &           & 0.617 & 0.521 & 0.489 & \textbf{0.207} & 0.385 & 0.092 & 0.311 & \textbf{0.057} \\
4 &          &         & \checkmark  & 0.611 & 0.478 & 0.472 & 0.187 & 0.327 & 0.079 & 0.279 & 0.056 \\
5 &          & \checkmark  & \checkmark  & 0.617 & 0.527 & 0.491 & 0.202 & 0.403 & \textbf{0.098} & 0.301 & 0.056 \\
6 & \checkmark & \checkmark  &  & \textbf{0.623}  & 0.462 & 0.496 & 0.163 & 0.417 & 0.084 & 0.313 & 0.050 \\
7 & \checkmark  & \checkmark & \checkmark   & 0.610 & 0.461 & \textbf{0.498} & 0.171 & \textbf{0.421} & 0.087 & \textbf{0.314} & 0.049 \\
\bottomrule
\vspace{-10pt}
\end{tabular}
\end{table}

\textbf{The performance gain of each proposed component.} We ablate the three key components described in Section~\ref{sec: components}, and show the results in Table~\ref{tab: component ablation}. Comparing rows 2, 3, and 4 with row 1 (VGM), we can see that the LTM brings the largest improvement over the baseline at the 2, 3, and 4-goal levels among the three components. For example, the LTM brings an increase in PR by \textbf{0.056} at the 3-goal level. Although applying the forgetting module only achieves improvements at the 2 and 4-goal levels, its cooperation with the LTM witnesses a significant increase in the SR/PR at the 1, 2, and 3-goal levels. More importantly, compared with VGM (row 1), the synergy of the three components (row 7) increases the PR by \textbf{0.049} (\textbf{10.9\%}), \textbf{0.092} (\textbf{21.9\%}), \textbf{0.038} (\textbf{13.8\%}) at the 2, 3 and 4-goal levels respectively. These results demonstrate that the three components are helpful for solving long-time navigation tasks with multiple sequential goals.

\textbf{The importance of the LTM.} Table~\ref{tab: node processing} presents the results of ablation experiments that demonstrate the importance of the LTM (described in Section~\ref{sec: Global node}). The first ablation (row 2) is to replace the feature in the LTM with that of a randomly selected node in the STM once the GATv2 encoding is finished so that the LTM is disabled. Row 2 shows that the agent's performance decreases at all difficulty levels compared with the full model (row 1). Furthermore, when the LTM feature is not incorporated into the WM (row 3), the performance also deteriorates. In summary, the LTM stores a scene-level feature essential for improving the success rate.

\begin{table}[]
\scriptsize
\caption{Ablation study of the LTM.}
  \label{tab: node processing}
\centering
\begin{tabular}{@{}cccccccccc@{}}
\toprule
\multirow{2}{*}{} & \multirow{2}{*}{Methods} & \multicolumn{2}{c}{\textbf{1-goal}} & \multicolumn{2}{c}{\textbf{2-goal}} & \multicolumn{2}{c}{\textbf{3-goal}} & \multicolumn{2}{c}{\textbf{4-goal}} \\
\cmidrule(lr){3-4} \cmidrule(lr){5-6} \cmidrule(lr){7-8} \cmidrule(lr){9-10}
 &   & \textbf{SR} & \textbf{SPL} & \textbf{PR} & \textbf{PPL} & \textbf{PR} & \textbf{PPL} & \textbf{PR} & \textbf{PPL} \\ \midrule

1 & Ours                      & 0.610 & 0.461 & 0.498 & 0.171 & 0.421 & 0.087 & 0.314 & 0.049 \\
2 & Ours w. random replace    & 0.602 & 0.365 & 0.464 & 0.122 & 0.373 & 0.065 & 0.306 & 0.046 \\
3 & Ours w.o. decoding LTM   & 0.604 & 0.456 & 0.492 & 0.169 & 0.393 & 0.083 & 0.309 & 0.048 \\

\bottomrule
\vspace{-10pt}
\end{tabular}
\end{table}

\begin{figure}
\centering
\subfigure[SPL/PPL v.s. $p$]{\label{fig:SPL vs p}
 \includegraphics[scale=0.185]{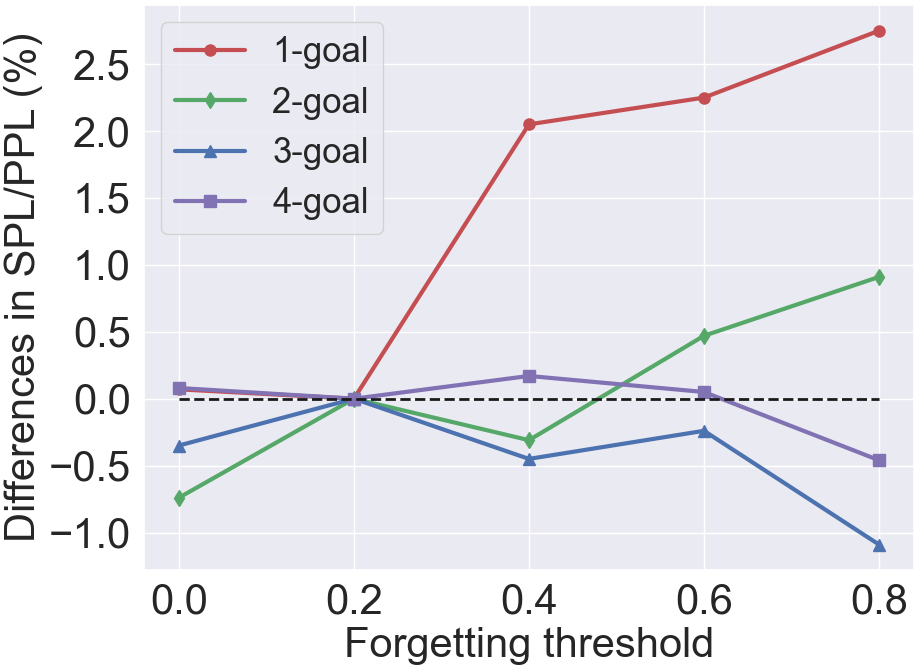}}
\subfigure[SR/PR v.s. $p$]{\label{fig:SR vs p}
 \includegraphics[scale=0.185]{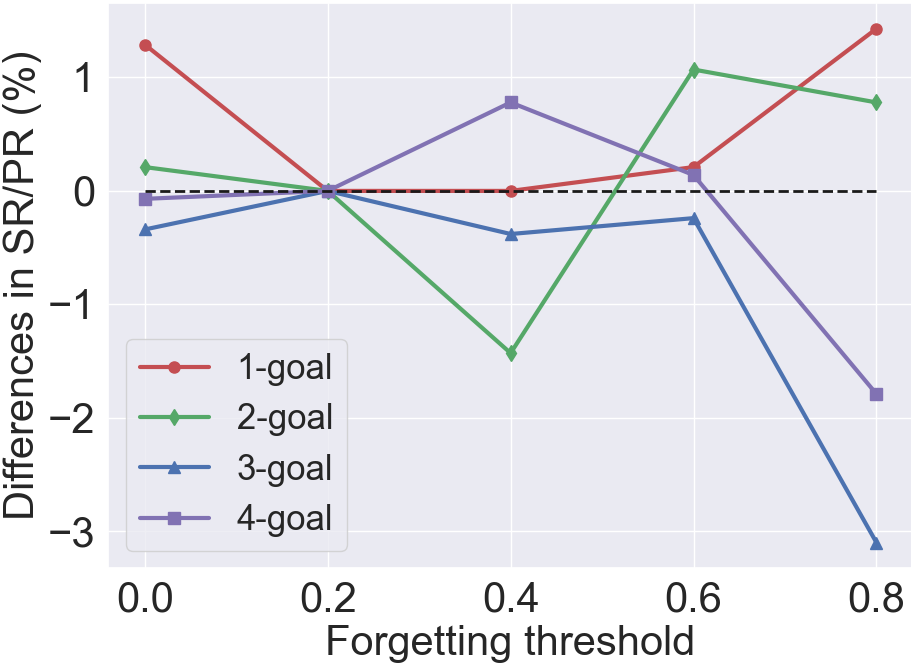}}
\subfigure[Average steps v.s. $p$]{\label{fig:Steps vs p}
 \includegraphics[scale=0.185]{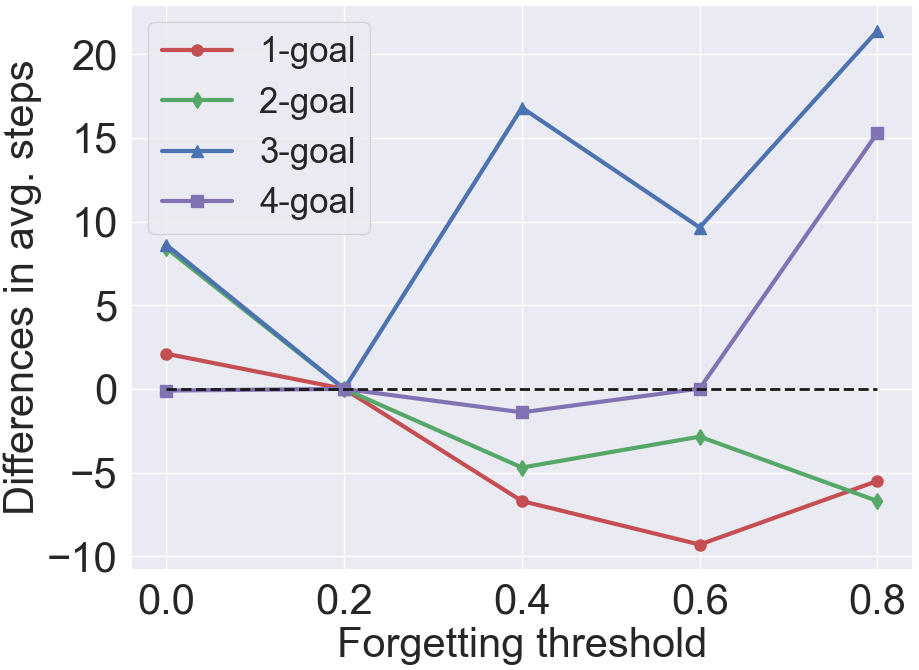}}
\caption{The agent's navigation performance at different forgetting threshold $p$.}
\label{fig:forgetting}
\end{figure}
\textbf{The correlation between the difficulty level and forgetting threshold}. We evaluate our model with different forgetting thresholds $p$ used by the WM (see Section~\ref{sec: forgetting mechanism}). The results are shown in Figure~\ref{fig:forgetting}. For clarity, the figure shows the performance differences between our default model (row 7 in Table~\ref{tab: component ablation}) and the variants. The four levels exhibit different trends: when $p$ increases (i.e., a larger fraction of STM is not incorporated into the WM), our model first witnesses increases and then drops in the SR/PR and SPL/PPL at the 3 and 4-goal levels while enjoying slight gains in these metrics at the 1 and 2-goal levels. At the 1-goal level, our model obtains the highest SR and SPL when $p=80\%$, which means that a large fraction of node features in the map are useless when the navigation task is easy. A similar trend can also be seen at the 2-goal level. In contrast, our model exhibits an increase in the PR (from \textbf{0.313} to \textbf{0.321}) and PPL (from \textbf{0.049} to \textbf{0.050}) at the 4-goal level when $p$ rises from 0\% to 40\%. However, if $p$ rises to 80\%, the two metrics see a precipitous decline, and the agent takes more than 20 steps to complete the tasks. These results suggest that excluding unimportant STM from the WM improves navigation performance in multi-goal navigation tasks. However, excluding an excessive fraction forces the agent not to utilize what it has explored, thus reducing the navigation efficiency.\par
\begin{figure}
  \centering
  \includegraphics[width=0.8\linewidth]{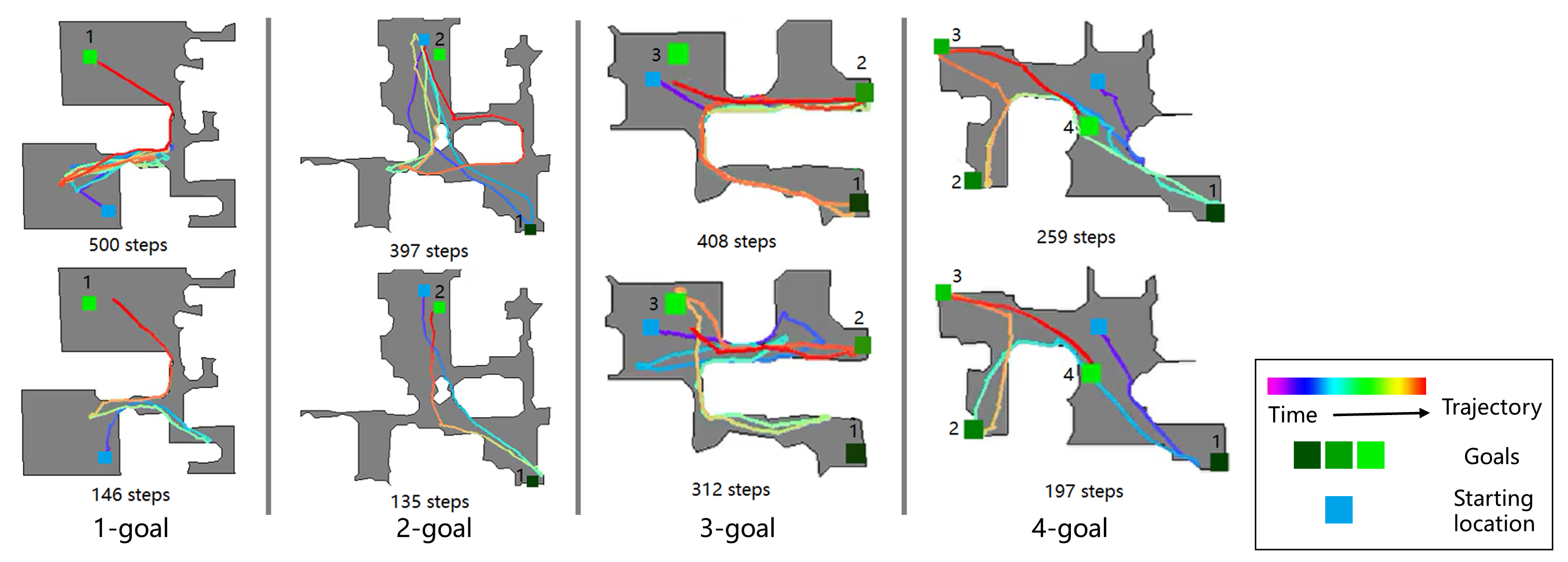}
  \caption{\textbf{Visualization of example episodes.} we compare selected episodes of the VGM (top row) and our MemoNav (bottom row) at four difficulty levels and visualize the top-down views. The number of navigation steps are shown at the bottom of each top-down view. Best viewed in color.}  
  \label{fig:vis}
\end{figure}

\subsection{Visualization Results}
To observe how the MemoNav improves the navigation performance, we show example episodes of VGM and our model at the four difficulty levels in Fig~\ref{fig:vis} (more examples are provided in the Appendix). We can see that the MemoNav agent takes fewer steps and its trajectories are smoother. VGM tends to spend a large proportion of time going in circles in narrow pathways. For instance, the 1st and 3rd columns in Figure~\ref{fig:vis} show that the VGM agent is trapped in a bottleneck connecting two rooms, while our agent uses the time steps wasted by VGM to efficiently explore the scene. This comparison shows that the MemoNav is more capable of avoiding deadlock. \par
\begin{figure}
  \centering
  \includegraphics[scale=0.4]{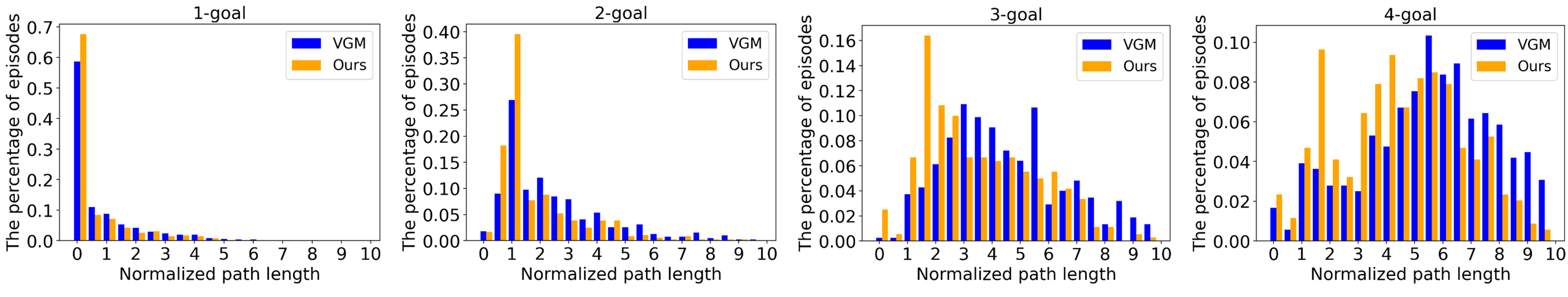}
  \caption{\textbf{Histogram of normalized path lengths of successful episodes for VGM \cite{kwon2021visual} and our model.} Our model has more short trajectories while VGM has a larger proportion of long episodes. }  
  \label{fig:stephist}
  \vspace{-10pt}
\end{figure}
We next draw a histogram of the normalized path lengths of successful episodes for VGM and our model to further investigate to what degree our model avoids deadlock. The normalized path length is calculated as $l_{norm} = l_{i} / \max \left(p_{i}, l_{i}\right) - 1$ where $p_{i}$ and $l_{i}$ are defined in Section~\ref{sec: eval metrics}. $l_{norm}$ indicates how many extra steps the agent takes compared to the shortest path. A larger $l_{norm}$ represents a less efficient navigation episode. Figure~\ref{fig:stephist} shows that the histogram of our model exhibits a less heavier-tailed distribution, especially on the 3-goal dataset. This distribution means that our model has fewer episodes with an extremely large number of steps. This result agrees with the visualization in Figure~\ref{fig:vis} , as it suggests that our model utilizes goal-relevant navigation history stored in the WM to reduce the number of wasted steps.\par
Analyzing these results, the reason why our model helps the agent escape from loops is twofold. Firstly, the WM excludes misleading node features and reduces the interference of irrelevant past observations. Secondly, the LTM provides the agent with a scene-level context that helps it locate the target from a global view.\par

\section{Conclusion}
This paper proposes MemoNav, a memory mechanism for ImageNav. This model flexibly retains the informative fraction of the short-term navigation memory via a forgetting module. We also add an extra global node to the topological map as a long-term memory that aggregates features in the short-term memory. The retained short-term memory and the long-term memory form the working memory that is used for generating action, Our model is only trained on a 1-goal dataset. Nevertheless, it is capable of solving a multi-goal navigation task. The results show that the MemoNav outperforms the baselines in 1-goal and multi-goal tasks and is more capable of escaping from deadlock.

\section*{Acknowledgements}
We would like to acknowledge Yuxi Wang, Yuqi Wang, Lue Fan, He Guan, Jiawei He, Zeyu Wang and the anonymous reviewers for their help, comments and suggestions.

\section*{Licenses for Referenced Datasets}
Gibson: \url{http://svl.stanford.edu/gibson2/assets/GDS_agreement.pdf}

\bibliography{references}

\section*{Checklist}

\begin{enumerate}

\item For all authors...
\begin{enumerate}
  \item Do the main claims made in the abstract and introduction accurately reflect the paper's contributions and scope? \answerYes{See Figure~\ref{fig:task_comparison} right and Figure`\ref{fig:model_overview}.}
  \item Did you describe the limitations of your work?
    \answerYes{Refer to Section \ref{sec: ablation} for the discussion about how the forgetting module probably undermines navigation efficiency. Additional description of the limitations of the proposed model is placed in the supplementary materials.}
  \item Did you discuss any potential negative societal impacts of your work?
    \answerNA{}
  \item Have you read the ethics review guidelines and ensured that your paper conforms to them?
    \answerYes{}
\end{enumerate}

\item If you are including theoretical results...
\begin{enumerate}
  \item Did you state the full set of assumptions of all theoretical results?
    \answerNA{}
        \item Did you include complete proofs of all theoretical results?
    \answerNA{}
\end{enumerate}

\item If you ran experiments...
\begin{enumerate}
  \item Did you include the code, data, and instructions needed to reproduce the main experimental results (either in the supplemental material or as a URL)?
    \answerYes{Refer to the supplementary materials for detailed instructions of how to reproduce the results, which include hyperparameters, implementation details and data generation process. }
  \item Did you specify all the training details (e.g., data splits, hyperparameters, how they were chosen)?
    \answerYes{}
        \item Did you report error bars (e.g., with respect to the random seed after running experiments multiple times)?
    \answerNo{We decided not to report error bars with respect to multiple random seeds, since multiple experiments on training the proposed model and the baselines exhibit large performance diversity. In addition, training the proposed model and the baselines is time-consuming and expensive.}
        \item Did you include the total amount of compute and the type of resources used (e.g., type of GPUs, internal cluster, or cloud provider)?
    \answerYes{See the supplementary materials.}
\end{enumerate}

\item If you are using existing assets (e.g., code, data, models) or curating/releasing new assets...
\begin{enumerate}
  \item If your work uses existing assets, did you cite the creators?
    \answerYes{}
  \item Did you mention the license of the assets?
    \answerYes{}
  \item Did you include any new assets either in the supplemental material or as a URL?
    \answerNA{}
  \item Did you discuss whether and how consent was obtained from people whose data you're using/curating?
    \answerNA{}
  \item Did you discuss whether the data you are using/curating contains personally identifiable information or offensive content?
    \answerNA{}
\end{enumerate}

\item If you used crowdsourcing or conducted research with human subjects...
\begin{enumerate}
  \item Did you include the full text of instructions given to participants and screenshots, if applicable?
    \answerNA{}
  \item Did you describe any potential participant risks, with links to Institutional Review Board (IRB) approvals, if applicable?
    \answerNA{}
  \item Did you include the estimated hourly wage paid to participants and the total amount spent on participant compensation?
    \answerNA{}
\end{enumerate}

\end{enumerate}


\appendix

\section{Appendix}
\subsection{Open-sourced Code}
We have open-sourced the implementation of the MemoNav, which can be found at: \url{*} 

\subsection{Limitations}
While the MemoNav witnesses a large improvement in the navigation success rate in multi-goal navigation tasks, it still encounters limitations. The proposed forgetting module is a post-processing method, as it has to obtain the attention scores of the decoder before deciding which nodes are to be forgotten. Future work can explore trainable forgetting modules. The second limitation is that our forgetting module does not reduce memory footprint, since the features of the forgotten nodes still exist in the map for localization. Moreover, the forgetting threshold in our experiments is fixed. Future work can merge our idea with Expire-span \cite{sukhbaatar2021not} to learn an adaptive forgetting threshold.

\subsection{Representative Models of Human Memory}
\begin{figure}
  \centering
  \includegraphics[scale=0.4]{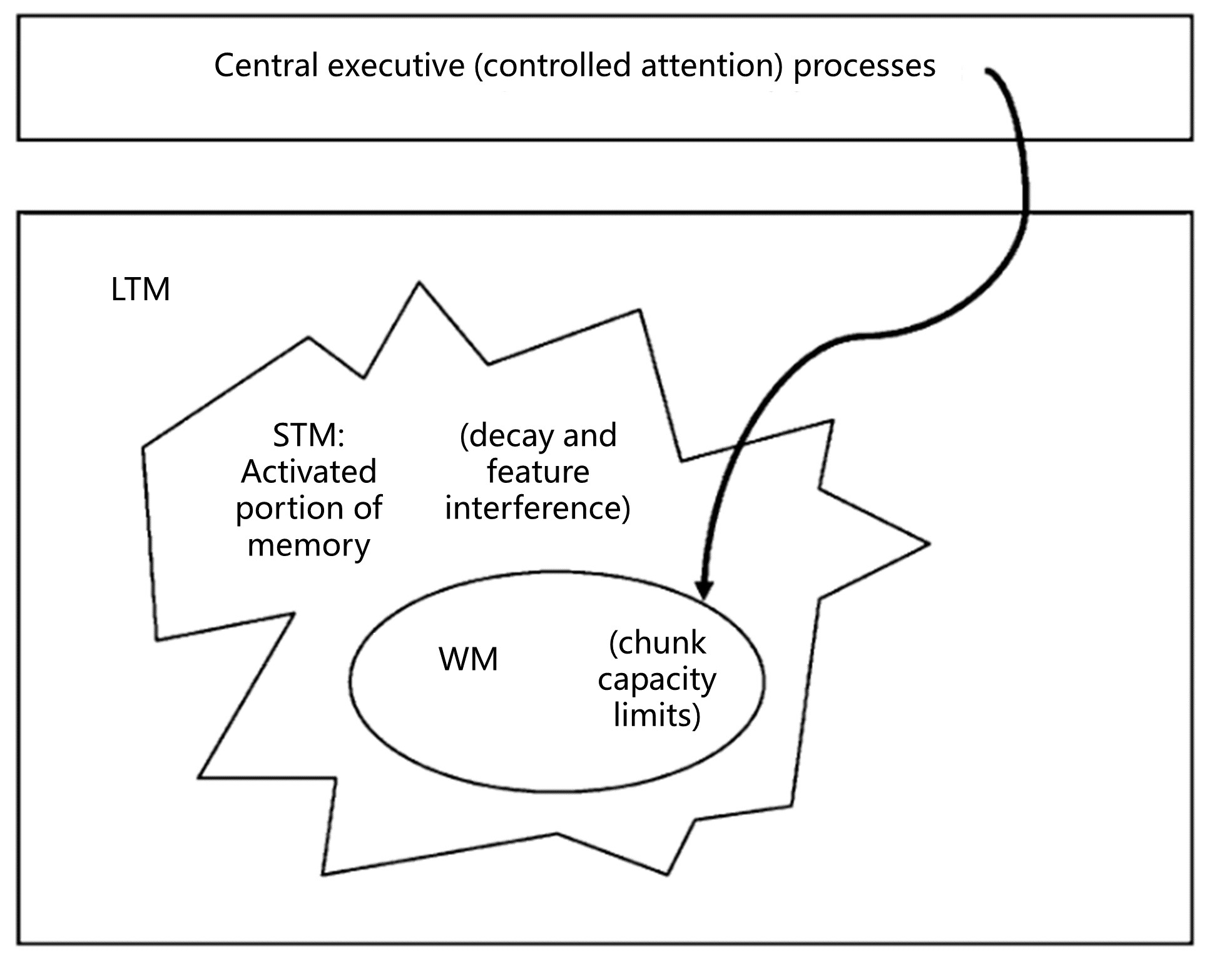}
  \caption{The memory model by Cowan et al. \cite{cowan2008differences}. This figure is borrowed and adapted from its original paper.}  
  \label{fig: cowan}
\end{figure}

Cowan et al. \cite{cowan2008differences} proposed a typical model describing the relationships among long-term memory (LTM), short-term memory (STM), and working memory (WM) of the human brain. According to their definitions, LTM is a large knowledge base and a record of prior experience; STM reflects faculties of the human mind that hold a limited amount of information in a very accessible state temporarily; WM includes part of STM and other processing mechanisms that help to utilize STM. Cowan et al. designed a framework depicting how WM is formed from STM and LTM (shown in Figure~\ref{fig: cowan}). This framework demonstrates that STM is derived from a temporarily activated subset of LTM. This activated subset may decay as a function of time unless it is refreshed. The useful fraction of STM is incorporated into WM via an attention mechanism to avoid misleading distractions. Subsequent work by Baddeley et al. \cite{baddeley2012working} suggests that the central executive manipulates memory by incorporating not only part of STM but also part of LTM to assist in making a decision.\par
We draw inspirations from the work by Cowan et al. \cite{cowan2008differences} and Baddeley et al. \cite{baddeley2012working} and reformulate the agent's navigation experience as the three types of memory defined above. The agent's current observations are analogous to STM since they are temporarily stored in the topological map; the global node aggregates graph node features and saves navigation history, thereby acting as LTM; the forgetting module retains the goal-relevant fraction of STM and incorporates the retained STM and LTM into WM, which is then input to the policy network.

\subsection{Implementation Details}
\subsubsection{Model Architecture}

\begin{wraptable}{r}{7cm}
\scriptsize
\caption{Network architecture of our model}
\vspace{10pt}
\label{tab: model param}
\centering
\begin{tabular}{@{}ccc@{}}
\toprule
Param & Meaning                                      & Value    \\ \midrule
$\mathcal{F}_{loc}$   & The pre-trained encoder used for  map update & ResNet18 \\
$d$     & The dimension of node feature                 & 512      \\
$\mathcal{F}_{enc}$ & \begin{tabular}[c]{@{}c@{}}The trainable encoder for embedding\\ observation and target images\end{tabular} & ResNet18 \\
$L$     & The  number of GATv2 layers                  & 3        \\
$H$     & The number of decoder heads                  & 4        \\ \bottomrule
\end{tabular}
\end{wraptable}

The structure of the memory encoding and decoding module (Figure 2(e) in the main paper) in the MemoNav remains the same as in the VGM \cite{kwon2021visual} and is shown in Figure~\ref{fig: encdec}. We maintain the module hyper-parameters specified in the supplementary of the VGM paper and list them in Table~\ref{tab: model param} for convenience.

\begin{figure}
  \centering
  \includegraphics[scale=0.4]{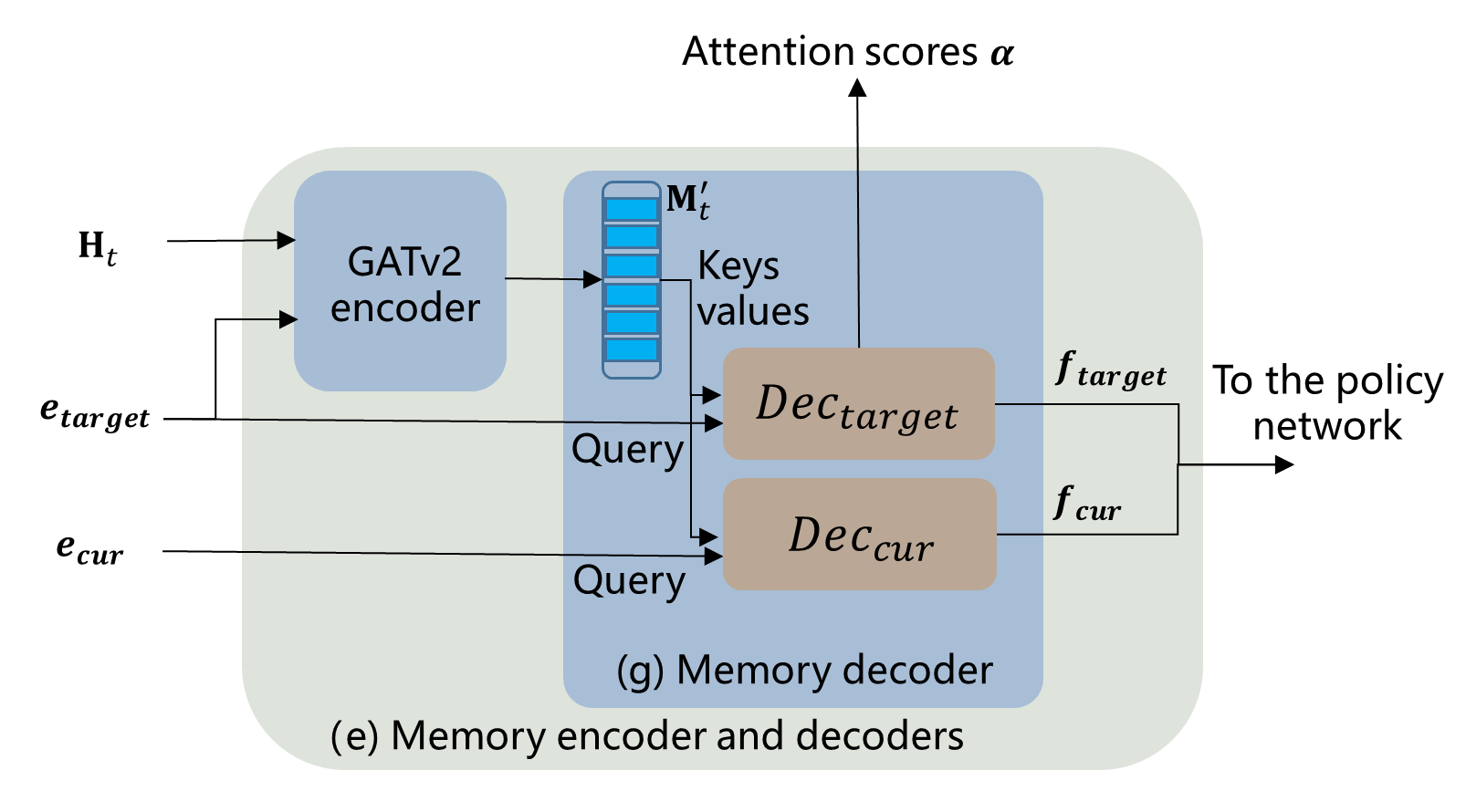}
  \caption{The structure of the memory encoding and decoding module displayed in Figure 2 of the main paper. $\mathbf{H}_t$ and $\mathbf{M}_t^{\prime}$ are the working memory before and after encoding. The attention scores $\mathbf{\alpha}$ generated in $\mathcal Dec_{target}$ are used in the forgetting module to retain informative STM.}  
  \label{fig: encdec}
\end{figure}

\subsubsection{Implementation of the MemoNav}
The forgetting module on the MemoNav requires the attention scores generated in the decoder $\mathcal Dec_{target}$. Therefore, our model needs to calculate the whole navigation pipeline before deciding which fraction of the STM should be retained. This lag means that the retained STM is incorporated into the WM at the next time step. The pseudo-code of the MemoNav is shown in Algorithm~\ref{alg: MemoNav}\par

\begin{algorithm}
\label{alg: MemoNav}
\caption{The implementation of the MemoNav}\label{algorithm}
\KwData{Empty topological map $G=\left\{V,E\right\}$, target image $\mathbf{I}_{target}$, current time step $t$, forgetting percentage $p$, trainable observation encoder $\mathcal F_{enc}$,  GATv2-based encoder $\operatorname{GATv2}$, Transformer decoders $\mathcal Dec_{target}$ and $\mathcal Dec_{cur}$, LSTM-based policy network $\operatorname{LSTM}$}
\KwResult{Navigation action $a_{t}$}
Long-term memory $n_{global} \leftarrow \mathbf{0} \in \mathbb{R}^d$\;
Attention scores for graph nodes $V$: $\mathbf{\alpha} \leftarrow \mathbf{0} \in \mathbb{R}^{\lvert V \rvert}$\;
\While{not $\operatorname{AgentCallStop}\left(\right)$}{
\tcp{Step 1: Update the topological map}
$\mathbf{I}_t \leftarrow \operatorname{GetCurrentPanorama}()$\;
$G.\operatorname{UpdateMap}\left(\mathbf{I}_t\right)$\; 

\tcp{Step 2: Retain the informative fraction of the STM}
Forgotten number $n \leftarrow \operatorname{Floor}\left(p \cdot \lvert V \rvert \right)$\;
Sorted indices $i \leftarrow \operatorname{Argsort}\left(\mathbf{\alpha}\right)$\;

Forgotten indices $i_{for} \leftarrow i\left[0\colon n\right]$\;
$G.\operatorname{RemoveNodes}\left(i_{for}\right)$\;

\tcp{Step 3: Memory encoding and decoding}
$\mathbf{V}\in\mathbb{R}^{\lvert V \rvert \times d} \leftarrow G.\operatorname{GetNodeFeatures}\left(\right)$\;
Encoded memory $ \mathbf{M}^{\prime} \leftarrow \operatorname{GATv2} \left(\left\{\mathbf{V}, n_{global}\right\}\right)$\;
$e_{cur} \leftarrow\mathcal{F}_{enc}(\mathbf{I}_{t}),\ e_{target} \leftarrow \mathcal{F}_{enc}(\mathbf{I}_{target})$\;
$f_{cur} \leftarrow \mathcal Dec_{cur}\left(e_{cur},\mathbf{M}^{\prime} \right),\  f_{target} \leftarrow \mathcal Dec_{target}\left(e_{target},\mathbf{M}^{\prime} \right)$\;
$\mathbf{\alpha} \leftarrow \mathcal Dec_{target}.\operatorname{GetAttScores}\left(\right)$

\tcp{Step 4: Action generation}
$x \leftarrow \operatorname{LSTM}\left(\operatorname{FC}\left(\left[f_{cur}, f_{target}, e_{cur}\right]\right)\right)$\;
$p\left(a_{t} \mid x\right)=\sigma\left(\operatorname{FC}\left(x\right)\right)$\;
$a_{t} \leftarrow \operatorname{SampleFromDistribution}\left(p\left(a_{t} \mid x\right)\right)$\;
}

\end{algorithm}

\subsubsection{Training Details}
We follow the two-step training routine and maintain the training hyper-parameters shown in Table 5 of the VGM paper \cite{kwon2021visual}. Our model is trained using imitation learning for 20k steps. Afterward, we finetune our model using PPO \cite{schulman2017proximal} for 10M steps. Due to the performance fluctuation intrinsic to reinforcement learning, the model at the 10M-th step is probably not the best. Therefore, we evaluate all checkpoints in the step range [9M, 10.4M] and select the best one. Our model compared with the baselines is the checkpoint at the 9.4M-th step.

\subsubsection{Details of Collecting Multi-goal Datasets}
\begin{figure}
  \centering
  \includegraphics[scale=0.4]{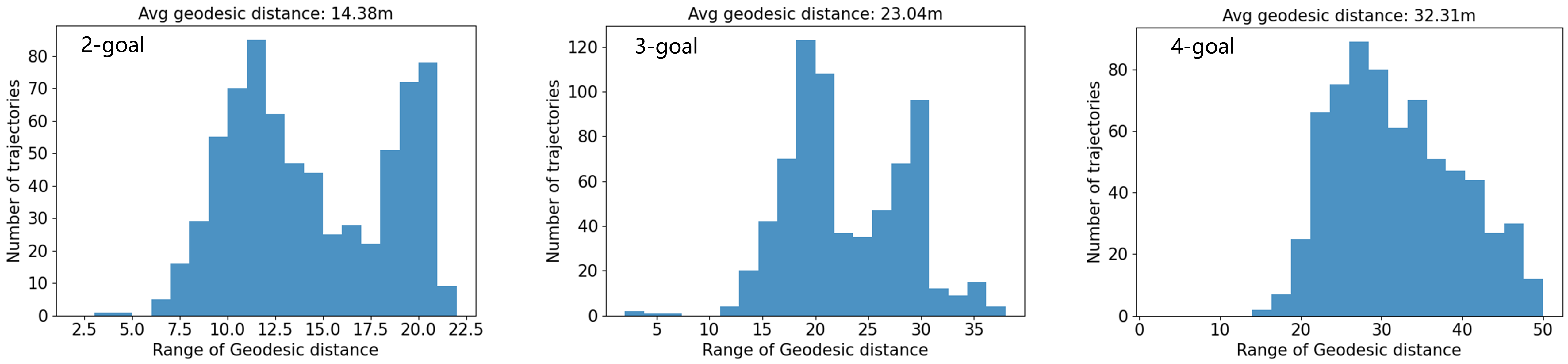}
  \caption{Histograms of geodesic distances for the multi-goal test datasets.}  
  \label{fig:geohist}
\end{figure}

We follow the format of the public 1-goal dataset in \cite{mezghani2021memory} and create 2-goal, 3goal, and 4-goal test datasets on the Gibson \cite{xia2018gibson} scene dataset. We generate 50 samples for each of the 14 test scenes. In each sample, we randomly choose target positions while still following five rules: (1) no obstacles appear inside a circle with a radius of 0.75 meters centered at each target; (2) the distance between two successive targets is no more than 10 meters; (3) all targets are placed on the same layer without altitude differences. (4) all targets are reachable from each other. (5) The final target is placed near a certain previous one with the distance between them smaller than 1.5 meters. The distributions of the total geodesic distances for the three difficulty levels are shown in Figure~\ref{fig:geohist}.

\subsubsection{Compute Requirements}
We utilize an RTX TITAN GPU for training and evaluating our models. The imitation learning phase takes 1.5 days to train while the reinforcement learning takes 5 days.\par
The computation in the GATv2-based encoder and the two Transformer decoders occupy the largest proportion of the run-time of the MemoNav. The computation complexity of the encoder and the decoders are $\mathcal{O}\left(\lvert V \rvert d^2 + \lvert E \rvert d\right)$ and $\mathcal{O} \left(\lvert V \rvert d\right)$, respectively. Using the forgetting module with a percentage threshold $p$, the computation complexity of the MemoNav can be flexibly decreased by reducing the number of nodes to $p\lvert V \rvert$.

\subsection{Extra Ablation Studies}
We conduct extra ablation experiments on the forgetting module in the MemoNav to further investigate how its design affects the agent's navigation performance. The results are shown in Table~\ref{tab: extra forgetting}.\par
\textbf{The origin of the attention scores used by the forgetting module}. The forgetting module in the MemoNav removes the uninformative STM according to the attention scores generated in $\mathcal Dec_{target}$, as described in Section 4.2.1. We change these scores to those generated in $\mathcal Dec_{cur}$. The result (row 2) shows that using the attention scores generated in $\mathcal Dec_{cur}$ as in $\mathcal Dec_{target}$, which means that the two decoders can both decide which fraction of the STM is informative. We choose $\mathcal Dec_{target}$ since it brings larger performance gains. \par
\textbf{The effectiveness of the forgetting module}. This ablation investigates whether it is effective to exclude the STM with attention scores ranking below the predefined percentage $p$. In this experiment,the forgetting module excludes a random fraction of the STM. We test this ablation model over five random seeds and report the average metrics. The result (row 3) shows that incorporating a random fraction of the STM into the WM leads to decreases in all metrics, which validates the effectiveness of our design for retaining informative STM. \par
\textbf{Training the MemoNav with the forgetting module}. As described in Section 4.2.1, the MemoNav is trained without the forgetting module. Here, we test the performance of the MemoNav trained with this module. The result (row 3) shows that training the MemoNav with the forgetting module leads to worse performance at the four difficulty levels.\par

\begin{table}[]
\scriptsize
\caption{Extra ablation studies of the forgetting module.}
  \label{tab: extra forgetting}
\centering
\begin{tabular}{@{}cccccccccc@{}}
\toprule
\multirow{2}{*}{} & \multirow{2}{*}{Methods} & \multicolumn{2}{c}{\textbf{1-goal}} & \multicolumn{2}{c}{\textbf{2-goal}} & \multicolumn{2}{c}{\textbf{3-goal}} & \multicolumn{2}{c}{\textbf{4-goal}} \\
\cmidrule(lr){3-4} \cmidrule(lr){5-6} \cmidrule(lr){7-8} \cmidrule(lr){9-10}
 &   & \textbf{SR} & \textbf{SPL} & \textbf{PR} & \textbf{PPL} & \textbf{PR} & \textbf{PPL} & \textbf{PR} & \textbf{PPL} \\ \midrule

1 & Ours                      & 0.610 & 0.461 & 0.498 & 0.171 & 0.421 & 0.087 & 0.314 & 0.049 \\
2 & Ours w. $\mathcal Dec_{cur}$ att. scores    & 0.616 & 0.453 & 0.513 & 0.175 & 0.413 & 0.084 & 0.323 & 0.051 \\
3 & Ours w. Random STM   & 0.597 & 0.446 & 0.479 & 0.153 & 0.391 & 0.081 & 0.307 & 0.048 \\
4 & Ours w. Training forgetting   & 0.601 & 0.416 & 0.462 & 0.132 & 0.375 & 0.068 & 0.266 & 0.039 \\

\bottomrule
\end{tabular}
\end{table}

\subsection{Extra Visualization Results}
\begin{figure}
\centering
\subfigure[]{
\begin{minipage}[t]{0.9\linewidth}
\label{fig: extra vis}
\centering
 \includegraphics[scale=0.7]{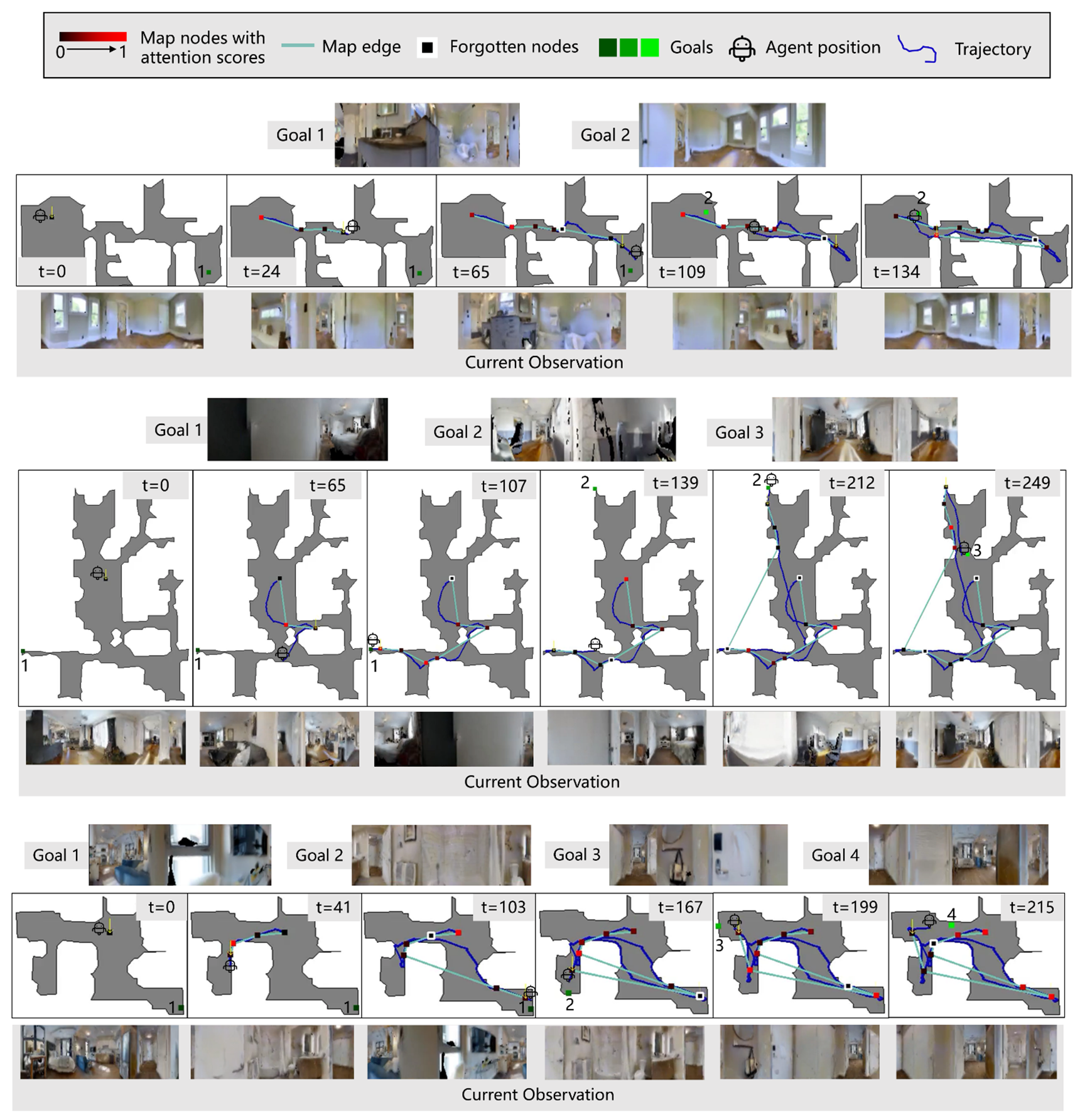}
 \end{minipage}
 }
\subfigure[]{
\begin{minipage}[t]{0.9\linewidth}
\label{fig: extra failure}
\centering
 \includegraphics[scale=0.35]{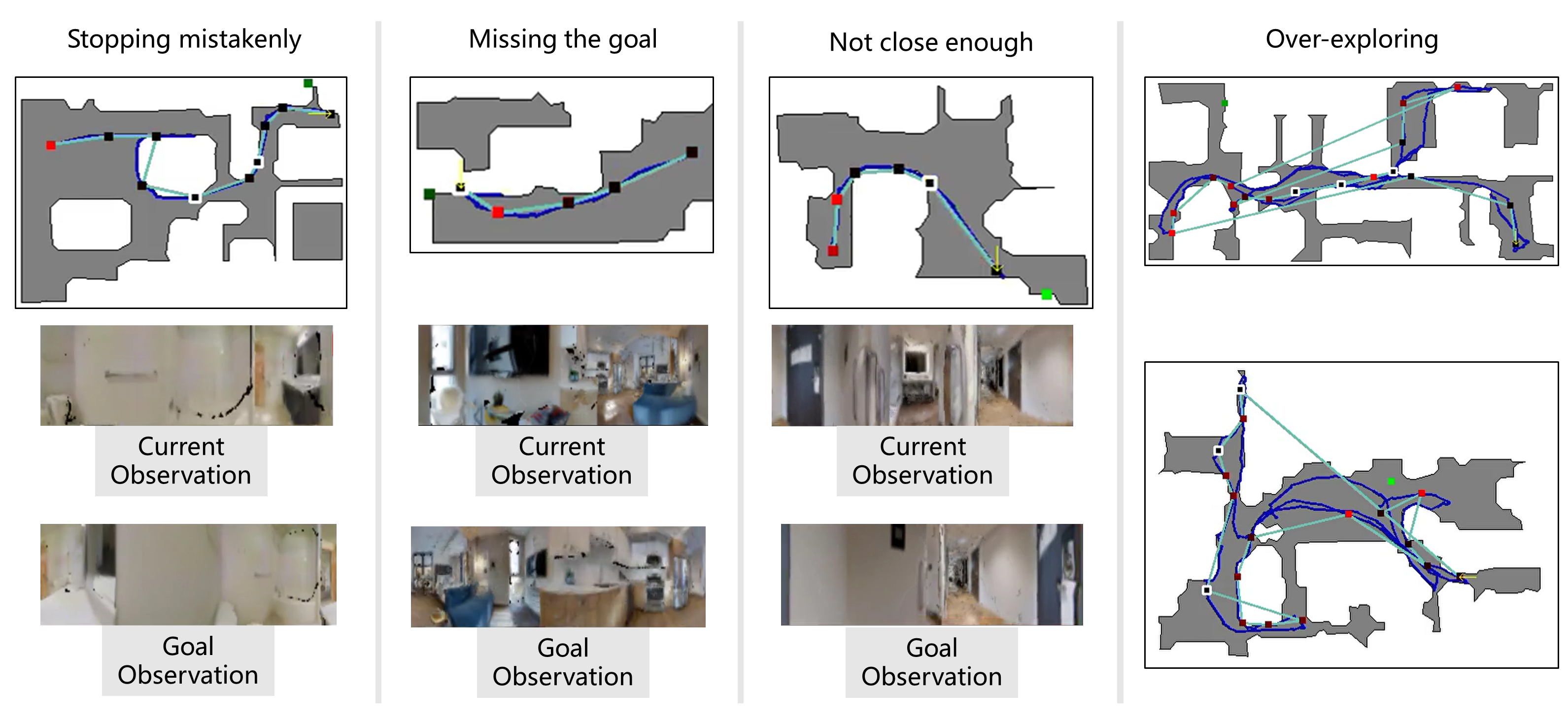}
 \end{minipage}
 }
\caption{(a) Examples of multi-goal navigation episodes. Each example shows both the topological map and the trajectory. The graph nodes are incrementally added to the map by the agent and selectively retained by the forgetting module in the MemoNav. The yellow downward arrow denotes the current localized node of the agent. (b) Examples of failed episodes. The agent encounters three major failure mode: \emph{Stopping mistakenly}, \emph{Missing the goal}, \emph{Not close enough}, and \emph{Over-exploring}.}
\label{fig: extra}
\end{figure}

We demonstrate more examples of multi-goal episodes in Figure~\ref{fig: extra vis}. The agent efficiently explores the scenes and finds sequential goals using the informative fraction of the node features in the map. These examples show that the MemoNav agent focuses high attention only on a small fraction of nodes and excludes nodes that are far away from the current goal. For example, in the 3-goal example, the agent forgets the topmost node when navigating to the 1st goal since this node is the farthest from the goal; the agent forgets the nodes at the bottom left corner when navigating to the 2nd and 3rd goals since these nodes are remote and uninformative. The comparison with the baseline for these examples is recorded in the supplementary videos. \par
We present examples of failed episodes in Figure~\ref{fig: extra failure} and record the proportions of various failure modes at all difficulty levels. The failure modes can mainly be categorized into four types: \emph{Stopping mistakenly}, \emph{Missing the goal}, \emph{Not close enough}, and \emph{Over-exploring}. The mode \emph{Stopping mistakenly} (34.2\%) means that the agent implements \verb+stop+ at the wrong place. The mode \emph{Missing the goal} (13.5\%) means that the agent has observed the goal but passes it. The mode \emph{Not close enough} (3.2\%) means that the agent attempts to reach the goal it sees but implements \verb+stop+ outside the successful range. The mode \emph{Over-exploring} (49.1\%) means that the agent spends too much time exploring open areas without any goals. The largest probability lies in \emph{Over-exploring} cases, most of which occur when the agent explores a large proportion of the scene but still fails to get close to the target area in a limited time.

\subsection{The Variation of the LTM}
\begin{figure}
  \centering
  \includegraphics[width=0.9\linewidth]{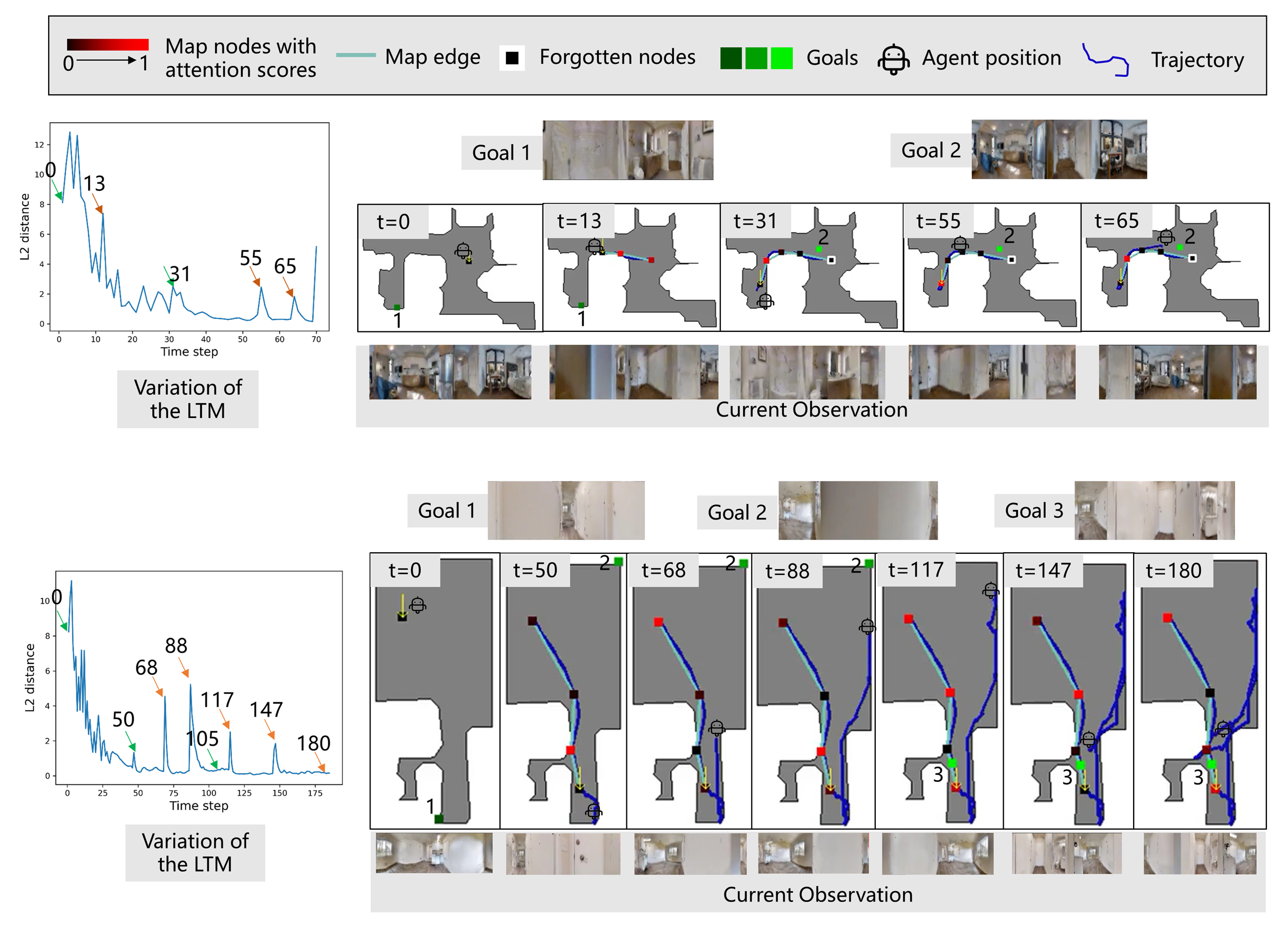}
  \caption{\textbf{Visualization of the LTM variation.} We show the agent's trajectories in two example episodes and visualize the agent's observations at the time steps when peaks appear on the LTM variation curves. The green arrows denote when the agent sets a new goal while the orange ones denote when peaks appear.}  
  \label{fig: LTM}
\end{figure}

To investigate how the feature in the LTM changes during navigation, we calculate the L2 distance between the features at every two successive time steps and show the curves for two examples in Figure~\ref{fig: LTM}. The two curves show similar trends: the L2 difference rapidly increases and then gradually converges to 0 while several peaks appear. To understand why the LTM variation shows such a trend, we visualize the agent's observations at the time steps of the peaks in Figure~\ref{fig: LTM}. Comparing these observations, we can see that the trend of the LTM variation exhibits two traits: (1) the LTM feature remains stable (the L2 difference is close to 0) if the agent is traveling in visited areas. For instance, in the 2-goal example (top row), the L2 difference steadily decreases in $t=31\sim 55$ during which the agent turns around and travels to visited areas; (2) the LTM feature significantly changes (a peak appears) if the agent captures novel views of the scene. For instance, in the 3-goal example (bottom row), the L2 difference curve exhibits peaks at $t=68$ when the agent passes a corner and at $t=88$ when the agent observes a novel open area. These results indicate that the LTM focuses higher attention on the agent's novel observations when aggregating the STM and stores the agent's exploration experience.

\end{document}